M. Abuella and B. Chowdhury, "Solar Power Forecasting Using Support Vector Regression," in *Proceedings of the American Society for Engineering Management International Annual Conference*, 2016.

# Solar Power Forecasting Using Support Vector Regression

Mohamed Abuella, Student Member, IEEE
Energy Production and Infrastructure Center
Department of Electrical and Computer Engineering
University of North Carolina at Charlotte
Charlotte, USA
Email: mabuella@uncc.edu

Badrul Chowdhury, Senior Member, IEEE
Energy Production and Infrastructure Center
Department of Electrical and Computer Engineering
University of North Carolina at Charlotte
Charlotte, USA
Email: b.chowdhury@uncc.edu

*Abstract*— Generation and load balance is required in the economic scheduling of generating units in the smart grid. Variable energy generations particularly from wind and solar energy resources are witnessing a rapid boost, and, it is anticipated that with a certain level of their penetration, they become noteworthy sources of uncertainty. As in the case of load demand, energy forecasting can also be used to mitigate some of the challenges that arise from the uncertainty in the resource. While wind energy forecasting research is considered mature, solar energy forecasting is witnessing a steadily growing attention from the research community. This paper presents a support vector regression model to produce solar power forecasts on a rolling basis for 24 hours ahead over an entire year, to mimic the practical business of energy forecasting. Twelve weather variables are brought from a high-quality benchmark dataset and new variables are extracted. The added value of heat index and wind speed as additional variables to the model is studied across different seasons. The support vector regression model's performance is compared with artificial neural networks, multiple linear regression models for energy forecasting.

I. INTRODUCTION

The wind and solar energy resources have created operational challenges for the electric power grid due to the uncertainty involved in their output in the short term. The intermittency of these resources may adversely affect the operation of the electric grid when the penetration level of these variable generations is high. Thus, wherever the variable generation resources are used, it becomes highly desirable to maintain higher than normal operating reserves and efficient energy storage systems to manage the power balance in the system. The operating reserves that use fossil fuel generating units should be kept as low as possible to get the highest benefit from the deployment of the variable generations. Therefore, the forecast of these renewable resources becomes a vital tool in the operation of the power systems and electricity markets [1].

As in wind power forecasting the solar power also consists of a variety of methods based on the time horizon being forecasted, the data available to the forecaster and the particular application of the forecast. The methods are broadly categorized according to the time horizon in which they generally show value. Methods that are common in solar forecasting include Numerical Weather Prediction (NWP) and Model Output Statistics (MOS) to produce forecasts, as well as hybrid techniques that combine ensemble forecasts, and Statistical Learning Methods [2]. In solar power forecasting up to 2-hours ahead, the most important input is the available observations of solar power, while for longer horizons NWP model becomes crucial for accurate forecasts [3]. Applying machine learning techniques directly to historical time-series of PV production associated with NWP outcomes have been placed among the top models in the last global competition of energy forecasting, GEFCom 2014 [4]. Moreover, the empirical study in [5] presents that errors of the solar power forecasting follow a non-parametric distribution, therefore, using non-parametric methods such as machine learning techniques can yield a better accuracy.

Support Vector Regression (SVR) is extended from support vector machines, a popular tool in machine learning. Time-series predictions by using SVM includes the load forecasting are explored in [6]. The SVM utilization in wind forecasting with severe ramp events is explained by a good extent in [7]. The prediction of solar irradiance from SVM and other machine learning methods are presented in [8], it concludes that SVM gives the best forecasts. The study [9] is for using SVR to forecast the solar power of a 1-MW PV power plant with the weather variables including the cloudiness, it shows that SVR and the cloudiness are improving the accuracy. The authors of [10] propose an approach including support vector machine and a weather classification methods to a PV-system of 20kW for a day ahead forecasting, the approach shows promising results. A benchmark study [11] for short-term wind and solar power forecasts over different sites in Europe, shows SVM models bring fairly good results. In [12] two separated models of SVR are constructed based on cloud cover to achieve more accurate forecasts of the solar irradiance.

Adding new variables to NWP models is not always valuable. By [13] found out that the accuracy of the solar irradiance forecasts based on aerosol chemical transport model depends on the sky conditions, whether it's clear or cloudy.

A recent review of a dozen of solar power forecasting studies are presented in [14], where the authors indicate that the majority of related publishing literature is focusing on solar irradiance instead of solar power due to the lack of the observed data of the solar power production. The solar irradiance forecasts need an additional step of a conversion to

**Note:** This is a pre-print of the full paper that published in *American Society for Engineering Management, International Annual Conference*, 2016, which can be referenced as below:
M. Abuella and B. Chowdhury, "Solar Power Forecasting Using Support Vector Regression," in *Proceedings of the American Society for Engineering Management 2016 International Annual Conference*, 2016.

the solar power and this might not be as accurate as the direct solar power forecasts.

## II. THE DATA

### A. Data Source

The data is derived from the Global Energy Forecasting Competition 2014 (GEFCom, 2014) which also including forecasting in the domains of electric load, wind power, solar power, and electricity prices. The solar data is for small solar power systems and their location is Australia [4].

### B. Data Description

The target variable is the solar power. There are 12 independent weather variables available from the European Centre for Medium-Range Weather Forecasts (ECMWF) that are used to produce solar forecasts. These are:

- Total column liquid water of cloud (tclw) - ($kg/m^2$).
- Total column ice water of cloud (tciw) - ($kg/m^2$)
- Surface pressure (SP) - (Pa).
- Relative humidity at 1000 mbar (r) - (%).
- Total cloud cover (TCC) - (0-1)
- 10 meter u-wind component (U) - (m/s).
- 10 meter v-wind component (V) - (m/s).
- 2 meter temperature (2T°) - (K)
- Surface solar radiation down (SSRD) - ($J/m^2$)-accumulated field.
- Surface thermal radiation down (STRD) - ($J/m^2$)-accumulated field.
- Top net solar radiation (TSR) - ($J/m^2$) -accumulated field
- Total precipitation (TP) - (m) - accumulated field.

The last four weather variables (i.e. solar and thermal radiations besides the precipitation) are given in accumulated field values, and not average values. They are increasing for every hour until the end of the day and then start again in accumulation. The wind variables are given as two components *u* and *v* representing wind directional components. The u-component of wind is positive for the west to east direction, while the v-component is positive for the south to north direction. The resultant vector of both wind components is the wind speed vector [15]. The polar form of wind speed ($|V|\angle\theta°$) in (1) is used later as an additional weather variable.

$$|V| = \sqrt{(u^2 + v^2)}, \qquad \theta = \tan^{-1}\frac{v}{u} \qquad (1)$$

Heat index is also an additional weather variable, which is extracted from the relative humidity and the air temperature variables. There are some methods to find the heat index, such as the regression method as follows [16].

Heat Index = -42.379 + 2.04901523T + 10.14333127R - 0.22475541TR - 6.83783*$10^{-3}T^2$ - 5.481717*$10^{-2}R^2$ + 1.22874*$10^{-3}T^2R$ + 8.5282*$10^{-4}TR^2$ - 1.99 *$10^{-6}T^2R^2$ (2)

Where: T is the air temperature (F°), and R is the relative humidity (percentage).

All available data are listed monthly as in Fig.1. a. The box plot of the annual distribution of the historical data of solar power is shown in Fig.1. b. Since the data from Australia, the distribution of the solar power over the year months obviously follows the southern hemisphere seasons, where the winter and summer seasons opposite to the northern hemisphere.

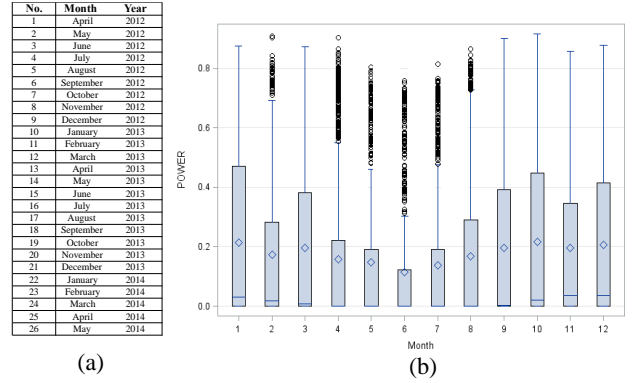

(a)        (b)

Fig.1. a) The entire available data. (b) Box plot of the distribution of observed solar power in 2012 months.

## III. SOLAR FORECAST MODELING

Machine Learning is the basic skeleton of the forecasting model in this paper, where the support vector regression (SVR) is deployed for short-term solar power forecasting. Fig.2 shows the flowchart of the machine learning, where the hypothesis block embodies the solar forecasting process, inputs (X) are the weather and past historical actual solar power data, and the outputs (Y) are the solar power forecasts. Whereas the training set is a part of the weather and the past actual solar power data, and the learning algorithm is SVR model.

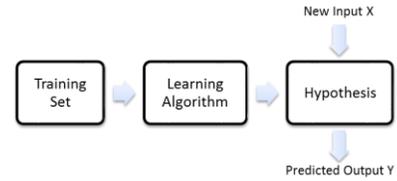

Fig.2. Machine Learning Flowchart

### A. Data Preparation

It is always a good idea to get the analysis of the used data before setting up the forecasting model. The historical data contains the actual solar power and 12 weather variables. The various steps of the data preparation are shown in Fig.3.

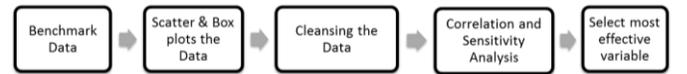

Fig.3. Flowchart diagram of data preparation

M. Abuella and B. Chowdhury, "Solar Power Forecasting Using Support Vector Regression," in *Proceedings of the American Society for Engineering Management International Annual Conference*, 2016.

*B. Variables Selection*

Selection of the predictor variables by plotting is cumbersome since there are many weather variables to choose from. So the greedy forward and backward search approaches besides the correlation analysis are carried out for the historical data to investigate the most effective variables. Fig.4 is the outcome of the variable selection. It is obvious that the solar irradiance, in addition to the time (hours), the surface irradiance and net top solar irradiance variables and their second order polynomial or quadratic terms have the highest correlation with solar power. The relative humidity and the air temperature also have a noticeable impact compared to other less effective variables.

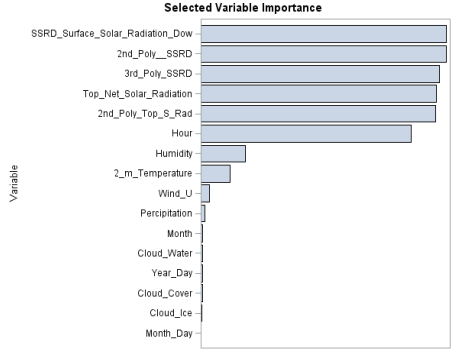

Fig.4. The order of most effective variable for solar power forecast

*C. SVR Model*

In general, support vector machine (SVM) is a machine-learning algorithm for classification applications. It's usually used with binary classes even if they are non-linearly separable in 2-D space, that is achieved by introducing a trick that transforms the classes into a higher dimensional space where the classes become linearly separable, and this trick is called kernel trick. The basic kernels are linear, polynomial, radial basis function [17]. Thereby, the support vector regression model SVR inherits some of the properties of the SVM. In SVR, the classification instead of being between the classes, it is basically a classification of the regression errors that are greater or less than a certain threshold as shown in Fig.5. For more details of SVR and its related mathematics refer to [18]. The main optimization and the kernel that are utilized with the SVR forecasting model here, are presented by (3) and (4).

The SVR requires the solution of the following optimization problem:

$$\min_{w,b,\xi} \frac{1}{2} W^T W + C \sum_{i=1}^{l} \xi_i \quad (3)$$

subject to $\quad y_i(W^T \phi(x_i) + b) \geq 1 - \xi_i$

Since $(x_i, y_i)$ training set pairs $i = 1, \ldots, l$. W is a normal unit vector that is perpendicular to the boundary margin, b is a slack variable, ε a threshold parameter, ξ deviation larger than ε. C > 0 is the penalty parameter. Training vectors $x_i$ are mapped into a higher dimensional space by the function $\phi$, this is the kernel trick $K(x_i, x_j) \equiv \phi(x_i)^T \phi(x_j)$.

For a kernel type of a Radial Basis Function (RBF):

$$K(x_i, x_j) = e^{-\gamma(\|x_i - x_j\|^2)}, \quad (4)$$

Where $\gamma$ (Gamma) is a kernel parameter.

*D. Grid Search*

The grid-search is a heuristic approach to find good SVR's hyper-parameters (C, γ), as they appear in (3), and (4). Many of methods are iterative processes, such as walking along a path until finding the optimal pair of these hyper-parameters. Since doing a complete grid-search may still be time-consuming, it's recommended to use the exponentially growing sequences of C and γ to identify good parameters (for example, $C = 2^{-5}, 2^{-3}, \ldots, 2^{15}$, $\gamma = 2^{-15}, 2^{-13}, \ldots, 2^{3}$) [19].

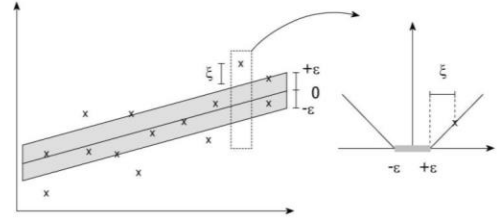

Fig.5. The Boundary margin and loss setting for a linear SVR [18]

III. MODEL RESULTS AND EVALUATION

The following measures are used to evaluate the accuracy of the forecasts and the model performance: plots and graphs, Root Mean Square Error (RMSE) are represented in (5), and a comparison with other models. For comparison purposes, the Multiple Linear Regression (MLR) Analysis model, and Artificial Neural Networks model (ANN) are used [20], [21]. The Library LIBSVM is used for simulation of SVR model [22].

$$RMSE = \sqrt{\frac{1}{n} \sum_{i=1}^{n} (Y_i - \hat{Y}_i)^2} \quad (5)$$

Where $\hat{Y}$ is the forecast of the solar power and $Y$ is the observed value of the solar power. $\hat{Y}$ and $Y$ are normalized values of the nominal power capacity of the solar power system. $n$ is the number of hours, it can be day hours or the month hours. The objective is to minimize RMSE for all forecasting hours to yield more accurate forecasts. If the training and testing of the model are carried out for just the daylight hours and filtering out the night hours (which have zero solar power generation), the RMSE should also be determined for these day hours only without including the night hours.

The grid search of the SVR's hyper-parameters is conducted to find out the optimal C and Gamma. In other words, searching for SVR's parameters that give a minimum RMSE. Fig.6 shows the RMSE contour, which results from the grid search of the SVR's hyper-parameters. The search grid is done by two cases: Firstly by using a cross-validation technique over arbitrary 30 days of the year and these resulted SVR's parameters are fixed and then used with all days [19]. While the other case, the adaptive hyper-parameters for each

**Note:** This is a pre-print of the full paper that published in *American Society for Engineering Management, International Annual Conference*, 2016, which can be referenced as below:
M. Abuella and B. Chowdhury, "Solar Power Forecasting Using Support Vector Regression," in *Proceedings of the American Society for Engineering Management 2016 International Annual Conference*, 2016.

month throughout the year are searched out and used in SVR to forecast the solar power of the days of each corresponding month adaptively.

TABLE I summarizes the evaluation results of the solar power forecasts of the daily forecasts on a rolling basis over a year. The evaluation (monthly-RMSE) of both cases, i.e., the fixed SVR's parameters and the adaptive parameters from June 2013 to May 2014 are presented. In February 2014 days, the adaptive SVR's hyper-parameters provide the best improvement 7% from using the fixed parameters case.

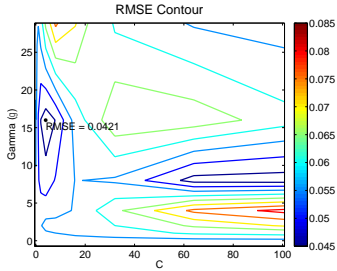

Fig.6. Grid Search of SVR's Hyper-Parameters, C, and Gamma

Thus, the SVR model is doing well even by using the fixed parameters (C=16, Gamma=1) comparing to the adaptive parameters case. Because the fixed parameters are validated by a cross-validation technique over random 30 days and seem this validation data is good enough to capture almost all information variation for the SVR model.

TABLE I
THE FORECASTS OF SVR MODEL OVER THE ENTIRE YEAR

| Month | Year | RMSE at C=16 & Gamma=1 | C | Gamma | RMSE at diff. C & Gamma | Imporvement (%) |
|---|---|---|---|---|---|---|
| June | 2013 | 0.0758 | 55 | 1.5 | 0.0734 | 3% |
| July | 2013 | 0.0872 | 100 | 1.0 | 0.0851 | 2% |
| August | 2013 | 0.0827 | 56 | 1.5 | 0.0818 | 1% |
| September | 2013 | 0.0751 | 16 | 1.0 | 0.0751 | 0% |
| October | 2013 | 0.0743 | 75 | 0.5 | 0.0712 | 4% |
| November | 2013 | 0.0670 | 75 | 0.5 | 0.0663 | 1% |
| December | 2013 | 0.0583 | 10 | 0.5 | 0.0574 | 2% |
| January | 2014 | 0.0557 | 99 | 1.4 | 0.0534 | 4% |
| February | 2014 | 0.0739 | 5 | 1.0 | 0.0684 | 7% |
| March | 2014 | 0.0817 | 11 | 1.0 | 0.0810 | 1% |
| April | 2014 | 0.0644 | 61 | 1.0 | 0.0635 | 1% |
| May | 2014 | 0.0553 | 16 | 1.0 | 0.0553 | 0% |

The added value of the additional weather variables, the heat index (HI) and wind speed, is investigated. These new variables are extracted from the available weather variables as in (1) and (2). The heat index is used with its two components variables; the humidity and the air temperature are also included. While for wind speed in the polar form, the wind components (u, v) are excluded from the weather data. The forecasts are generated for all days of the year, as shown in TABLE II. The forecasts without using heat index and wind speed are the similar of those in SVR's adaptive parameters case. Adding heat index and wind speed brings some improvement with September 2014 equals 4% and insignificant improvement in some months. Whereas in other months, adding these new variables reduces the accuracy of the model, as in its worst case with February 2014, the decline of the accuracy drops to 16% from the case of not including the heat index and wind speed. The main reason behind this is the seasonal trend of the weather data.

To see this seasonal variation in a clear picture, the correlation analysis is conducted between the actual solar power and six related weather variables, as shown in Fig.7. In this figure, the correlation results of three cases are presented. The first case is for training data of SVR, which includes all days from first day to the forecasted day, as shown in Fig.1. a. The other two cases of the correlation analysis are for February and September months' data when the most change of using heat index and wind speed occur. From Fig.7, it's obvious in September case; the correlation analysis is showing an almost similar trend of that in the training data of SVR model. While for February the trend of the heat index and wind speed variables is opposite of that in the training data of SVR model. Hence, in February 2014 and by including heat index and wind speed, SVR model suffers from a learning pattern change from its training data and that leads to the big reduction of forecasts accuracy.

TABLE II
THE RESULT OF USING HEAT INDEX AND WIND SPEED

| Month | Year | RMSE w/o HI & Wind | RMSE w. HI & Wind | RMSE Difference (%) |
|---|---|---|---|---|
| June | 2013 | 0.0734 | 0.0746 | -2% |
| July | 2013 | 0.0851 | 0.0845 | 1% |
| August | 2013 | 0.0818 | 0.0834 | -2% |
| September | 2013 | 0.0751 | 0.0722 | 4% |
| October | 2013 | 0.0712 | 0.0698 | 2% |
| November | 2013 | 0.0663 | 0.0665 | 0% |
| December | 2013 | 0.0574 | 0.0572 | 0% |
| January | 2014 | 0.0534 | 0.0562 | -5% |
| February | 2014 | 0.0684 | 0.0793 | -16% |
| March | 2014 | 0.0810 | 0.0846 | -4% |
| April | 2014 | 0.0635 | 0.0681 | -7% |
| May | 2014 | 0.0553 | 0.0549 | 1% |

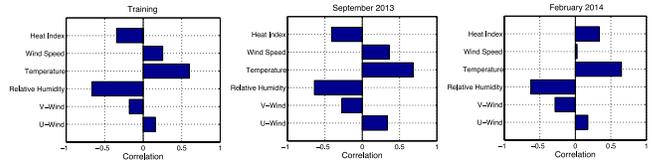

Fig.7. The solar power correlation plots with training and two months' data

The comparison of with other models is evaluated for SVR model in comparison with Multiple Linear Regression (MLR) and Neural Networks (ANN) models. For this purpose, the daily RMSE of last month days, May 2014 is shown in Fig.8. By the aggregation of daily RMSEs over the whole month, ANN and SVR models produce more accurate forecasts than MLR model. However, in some days MLR can have a lower RMSE as in 9th, 16th, 26th, and 30th days. The fluctuation in daily RMSE can be seen and it's a normal trend in the forecasting models. Since it appears that for a certain day a forecasting model could do a good job, but with another day the same model doesn't do well.

M. Abuella and B. Chowdhury, "Solar Power Forecasting Using Support Vector Regression," in *Proceedings of the American Society for Engineering Management International Annual Conference*, 2016.

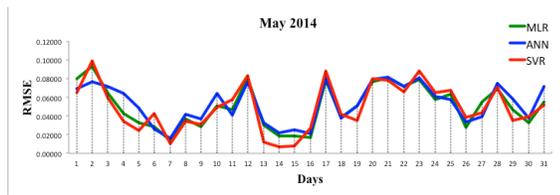

Fig.8. Daily RMSE of Different Models for May 2014

To get a broader evaluation of SVR's performance, the comparison with MLR and ANN models over a complete year is done as well, as shown in Fig.9. It is clear that ANN has lower monthly RMSEs. For some months SVR model can be as good as ANN. Therefore, SVR model has more accurate forecasts than MLR and less than ANN. The SVR has an advantage that cannot be shown illustratively, that it's more robust than ANN. Since SVR doesn't suffer from the local minima issue as ANN does. Due to the local minima issue in the fitting algorithm of ANN, thereby each time ANN runs it gives a different result for the same forecasting step (hour), so to overcome this, ANN forecasts are resulted from combining 10 forecasting values for each step [23].

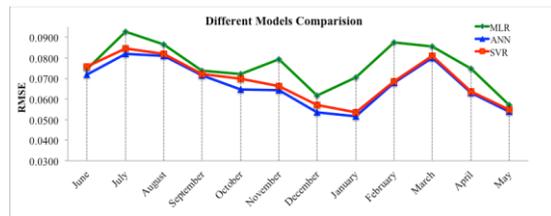

Fig.9. Monthly RMSE of Different Models

V. CONCLUSION

Whereas artificial neural networks suffer from multiple local minima, the support vector regression model's solution is global and unique. Furthermore, SVR is less prone to overfitting issue; these features make SVR more practical. Thereby, the SVR forecasting model is more accurate than multiple linear regression model and at the same time, it's more robust than ANN.

The added value study of including heat index and wind speed as additional weather variables leads to that these variables can improve the models' performance in some months. On the other hand, for other months that have a data pattern different from that in SVR training data, adding heat index and wind speed decreases the forecasts accuracy.

In the clear sky hours, the model produces more accurate forecasts than cloudy hours. The more accurate weather forecasts we use, the more accurate solar power forecasts will be produced. With additional historical data, the model performance will improve.

Although in a big picture, a particular forecasting model gives better forecasts, this forecasting model in some cases could not do well. Therefore, the combining models can be the efficient option to get the best from the different models, the authors are going to take the combining models into account in further work.

**Note:** This is a pre-print of the full paper that published in *American Society for Engineering Management, International Annual Conference*, 2016, which can be referenced as below:
M. Abuella and B. Chowdhury, "Solar Power Forecasting Using Support Vector Regression," in *Proceedings of the American Society for Engineering Management 2016 International Annual Conference*, 2016.